%% file: main.tex
\documentclass[10pt,twocolumn,letterpaper]{article}

\usepackage{wacv}              
\usepackage{algorithm}
\usepackage{algorithmic}
\usepackage{multirow}

\input{preamble}

\newcommand{\1}[1]{\mathbf{1}\!\left[#1\right]}

\definecolor{wacvblue}{rgb}{0.21,0.49,0.74}
\usepackage[pagebackref,breaklinks,colorlinks,allcolors=wacvblue]{hyperref}

\def\wacvPaperID{2} 
\def\confName{WACV}
\def\confYear{2026}

\title{Seeing Isn’t Believing: Context-Aware Adversarial Patch Synthesis via Conditional GAN}

\author{
Roie Kazoom*, Alon Goldberg, Hodaya Cohen, Ofer Hadar\\
Ben Gurion University of the Negev \quad
{\tt\small roieka@post.bgu.ac.il}
}

\begin{document}


\twocolumn[{
  \renewcommand\twocolumn[1][]{#1}
  \maketitle

  \begin{center}
    \includegraphics[width=0.9\textwidth]{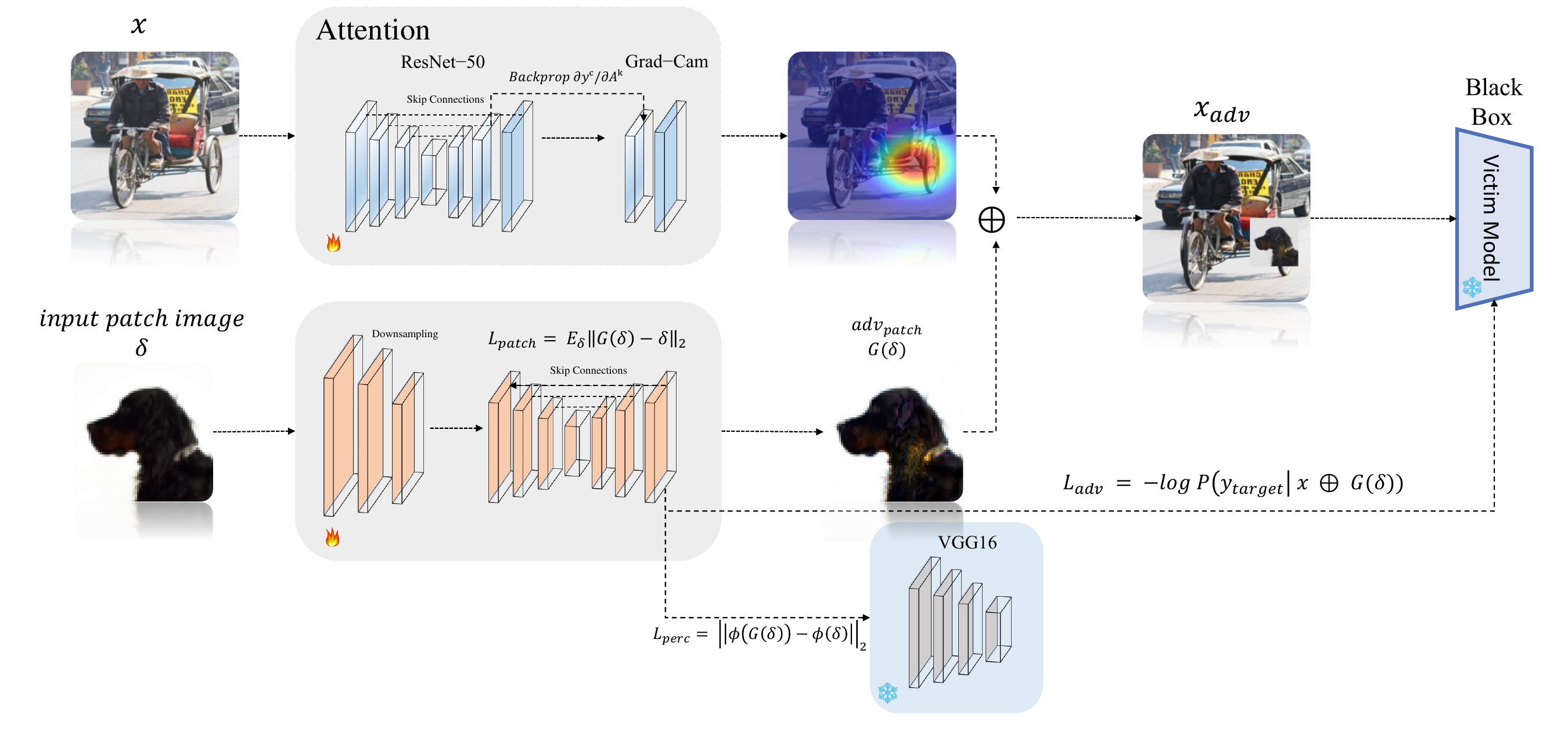}
    \captionof{figure}{Overall attack pipeline. Given an input image \(x\), we first extract Grad-CAM heatmaps from a surrogate ResNet-50 to localize semantically salient regions. A U-Net generator \(G\) consumes the seed patch \(\delta\) to synthesize an adversarial patch \(G(\delta)\). The patch is placed on \(x\) to form \(x_{\mathrm{adv}}\), which is then fed to the black-box victim model. We jointly optimize three losses: (1) adversarial loss \(L_{\mathrm{adv}}=-\log P(y_{\mathrm{target}}\mid x\oplus G(\delta))\), (2) pixel-level perceptual loss \(L_{\mathrm{patch}}=\mathbb{E}_{\delta}\|G(\delta)-\delta\|_{2}\), and (3) deep feature consistency loss \(L_{\mathrm{perc}}=\|\phi(G(\delta))-\phi(\delta)\|_{2}\) via a frozen VGG16.}

    \label{fig:attack_flow}
  \end{center}
}]

\begin{abstract}
Adversarial patch attacks pose a severe threat to deep neural networks, yet most existing approaches rely on unrealistic white-box assumptions, untargeted objectives, or produce visually conspicuous patches that limit real-world applicability. 
In this work, we introduce a novel framework for \textbf{fully controllable adversarial patch generation}, where the attacker can freely choose both the input image $x$ and the target class $y_{\text{target}}$, thereby dictating the exact misclassification outcome. 
Our method combines a generative U-Net design with \textbf{Grad-CAM-guided patch placement}, enabling semantic-aware localization that maximizes attack effectiveness while preserving visual realism. 
Extensive experiments across convolutional networks (DenseNet-121, ResNet-50) and vision transformers (ViT-B/16, Swin-B/16, among others) demonstrate that our approach achieves \textbf{state-of-the-art performance} across all settings, with attack success rates (ASR) and target-class success (TCS) consistently exceeding \textbf{99\%}. Importantly, we show that our method not only outperforms prior \textit{white-box} attacks and \textit{untargeted} baselines, but also surpasses existing \textit{non-realistic} approaches that produce detectable artifacts. 
By simultaneously ensuring realism, targeted control, and black-box applicability-the three most challenging dimensions of patch-based attacks-our framework establishes a new benchmark for adversarial robustness research, bridging the gap between theoretical attack strength and practical stealthiness. 
\end{abstract}

\section{Introduction}

\begin{figure*}[!t]
  \centering
  \includegraphics[width=0.7\textwidth]{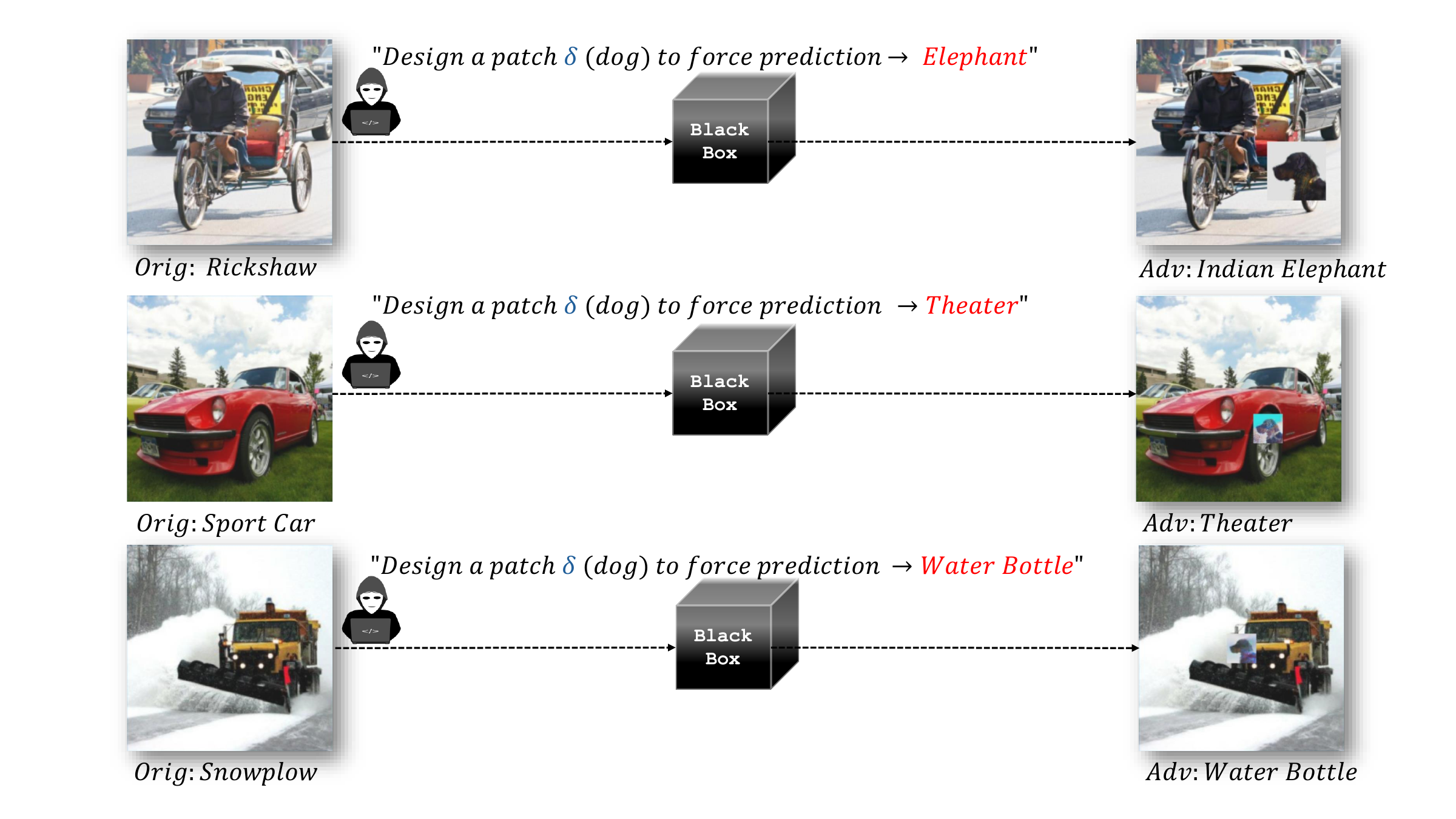}
  \caption{
Targeted adversarial patch attack framework. 
An input image $x$ is overlaid with an attacker-chosen patch $\delta$ (highlighted in \textcolor{blue}{blue}), 
producing an adversarial example $x \oplus \delta$. 
The adversarial input is passed to a black-box model, which is forced to predict an attacker-specified target class $y_{target}$ (highlighted in \textcolor{red}{red}). 
This figure emphasizes the two degrees of attacker control: (1) designing the adversarial patch $\delta$, 
and (2) selecting the desired misclassification target $y_{target}$. 
Arrows indicate the attack flow from clean input to adversarial output.}
  \label{fig:teaser_attacker}
\end{figure*}

Deep neural networks have revolutionized computer vision, achieving state-of-the-art accuracy on tasks such as image classification, object detection, and segmentation. However, they remain vulnerable to adversarial attacks-carefully crafted perturbations that can drastically alter model predictions while remaining imperceptible to human observers. This fragility poses serious risks in safety-critical domains like autonomous driving, medical imaging, and surveillance. Adversarial patch attacks form a particularly potent subclass: instead of small, distributed noise, they learn a localized pattern that can be printed and physically applied to real scenes. Brown \textit{et al.} first demonstrated the universal adversarial patch-a single overlay that consistently misleads classifiers across diverse inputs \cite{brown2017adversarial}. Eykholt \textit{et al.} extended this concept to object detection with the RP2 framework, showing that carefully placed “stickers” could fool YOLO models under real-world conditions \cite{eykholt2018robust, kazoom2025from}. Subsequent work has aimed to enhance patch realism, transferability, and robustness. Liu \textit{et al.} introduced the Perceptual-Sensitive GAN (PS-GAN), synthesizing visually coherent patches without sacrificing attack success rates \cite{liu2019perceptual}. Other approaches have incorporated physical constraints, spatial transformations, and environmental variations to ensure effectiveness outside the lab. Meanwhile, the shift from convolutional backbones to Vision Transformers (ViTs)-which process images as sequences of non-overlapping patches via self-attention-has spurred new investigations; Shao’s Random Position Adversarial Patch (G-Patch) employs a GAN-like generator to create universal patches for ViTs \cite{shao2023random}, achieving up to 97.1\% success on ViT-B/16.

In this work, we push adversarial patch research into the Transformer era with a \emph{targeted}, \emph{realism-aware} GAN framework under strict black-box constraints. Unlike prior methods that simply maximize misclassification, our generator consumes a real input image rather than random noise and produces a patch conditioned on an attacker-specified target class-enabling precise manipulation of the victim’s perception. To ensure both high attack success and visual plausibility, we jointly optimize three losses: (1) an adversarial loss that maximizes the victim’s predicted probability of the target class, (2) a pixel-level perceptual loss that preserves similarity to the seed patch, and (3) a deep feature consistency loss via a frozen VGG network to enforce semantic coherence. Crucially, we guide patch placement using Grad-CAM heatmaps extracted from a surrogate ResNet-50 \cite{selvaraju2017grad}, localizing perturbations to semantically salient regions without querying victim gradients. This purely black-box, attention-driven design yields strong generalization across both convolutional and Transformer classifiers, demonstrating consistent, high targeted attack success rates without any victim-model fine-tuning or additional gradient access.

\noindent\textbf{This paper makes the following contributions:}
\begin{enumerate}
  \item We propose a \emph{targeted}, realism-aware conditional GAN framework for adversarial patch synthesis that consumes a real input image and an attacker-specified target class, enabling precise control over the victim’s predicted label while maintaining visual plausibility.
  \item We introduce a purely black-box, attention-driven attack pipeline that leverages Grad-CAM heatmaps from a surrogate ResNet-50 to guide patch placement-requiring no gradient access to the victim model \cite{selvaraju2017grad}.
  
  \item We formulate a multi-objective loss combining (1) an adversarial loss to maximize the target‐class probability, (2) a pixel‐level perceptual loss to preserve seed‐patch similarity, and (3) a deep feature consistency loss via a frozen VGG network to enforce semantic coherence.
  \item We demonstrate strong generalization across diverse ImageNet-pretrained architectures-including both convolutional backbones and Vision Transformers-achieving state-of-the-art targeted attack success rates without any victim-model fine-tuning. Unlike prior patch attacks that address realism, black-box feasibility, or universality in isolation, our method uniquely unifies \textbf{image-conditioned patch synthesis}, \textbf{saliency-guided placement}, and \textbf{targeted misclassification} within a strict black-box setting.

\end{enumerate}

\section{Related Work}
Adversarial perturbations have been widely explored in the literature, ranging from training-free detection approaches~\cite{Kazoom2025DontLagRAG}, to robustness evaluation in natural language~\cite{Kazoom2025VAULT}, and defenses for object detection models~\cite{Kazoom2024EnhancingRobustness}. Adversarial patch attacks form a distinctive subclass of adversarial examples~\cite{szegedy2014intriguing}, where the perturbation is spatially localized rather than distributed across the entire input. 
Given an image $I \in \mathbb{R}^{H \times W \times 3}$, a binary mask $M \in \{0,1\}^{H \times W}$, and a patch pattern $P$, the perturbed image is constructed as
\begin{equation}
    I' = (1 - M) \odot I + M \odot P,
\end{equation}
where $\odot$ denotes element-wise multiplication. 
The design of $P$ determines both the success of the attack and its transferability across models.

\subsection{Adversarial Patch Attacks}

Adversarial patch attacks introduce localized, high-energy patterns applied directly to the image to manipulate model predictions. Unlike $\ell_p$-bounded perturbations, patches exploit spatial and semantic biases in modern vision models by inserting visually dominant signals into the scene. Their effectiveness makes them a practical threat model, particularly in settings where models must operate on natural images or real-world inputs.

\noindent\textbf{White-box patch attacks.}
In the white-box setting, the attacker has full access to the victim model's parameters, gradients, and logits. Patch optimization is performed via direct backpropagation through the victim model, solving
\begin{equation}
\theta^{*} = \arg\max_{\theta}\; \mathcal{L}_{\text{adv}}\big(f_{\text{victim}}(x \oplus P_{\theta}),\, y_{\text{target}}\big),
\end{equation}
where $P_{\theta}$ denotes the learnable patch. While this enables highly optimized attacks, full gradient access is often unrealistic in deployed or proprietary systems.

\noindent\textbf{Black-box patch attacks.}
Black-box attacks remove access to internal gradients or model parameters, allowing the attacker to observe only model outputs (e.g., softmax scores or top-$k$ labels). Without gradients, optimization typically follows either (1) query-based gradient approximation or (2) surrogate-based transfer, where a separate model is used to guide patch learning. A surrogate-based black-box attack therefore optimizes
\begin{equation}
\theta^{*} = \arg\max_{\theta}\; \mathcal{L}_{\text{adv}}\big(f_{\text{surrogate}}(x \oplus P_{\theta}),\, y_{\text{target}}\big),
\end{equation}
and directly applies the resulting patch to the victim model. Transferability of high-level patch features across architectures makes this strategy effective and practical for real-world threat scenarios.

\noindent\textbf{Universal and early patch attacks.}
Early works proposed universal adversarial patches~\cite{brown2017adversarial}, optimized to maximize expected misclassification under data distribution $\mathcal{D}$:
\begin{equation}
P^{*} = \arg\max_{P} \; \mathbb{E}_{x\sim\mathcal{D}}\big[\mathcal{L}(f(x\oplus P),\, y)\big].
\end{equation}
These patches demonstrated strong attacks but typically required white-box access and lacked semantic realism, limiting their robustness to transformations such as rotation or illumination changes~\cite{eykholt2018robust}. Later works explored vehicle- and scene-specific attacks~\cite{geng2023adversarial, shao2023random}, yet still relied on handcrafted or visually conspicuous textures.

\noindent\textbf{Targeted patch attacks.}
Targeted patch attacks aim to steer the model toward a specific target class $\tilde{y}$:
\begin{equation}
P^{*} = \arg\max_{P} \; \mathbb{E}_{x\sim\mathcal{D}}\big[ \log f_{\tilde{y}}(x \oplus P) \big].
\end{equation}
These approaches typically require careful optimization of semantic patterns while maintaining strong generalization across diverse images. Although targeted attacks have been demonstrated on large-scale models such as ViTs~\cite{shao2023random}, many lack realism or transfer poorly to black-box settings.

\noindent\textbf{Realistic patch generation.}
To improve stealthiness and reduce visual detectability, realism-aware generative models have been introduced. For example, PS-GAN~\cite{liu2019perceptual} balances adversarial loss with perceptual similarity:
\begin{equation}
\min_{G}\; \max_{D}\; \mathcal{L}_{\text{adv}} + \lambda\,\big\| P_{\theta} - I_{\text{ref}} \big\|_2^2,
\end{equation}
encouraging patches that resemble natural image content while still maximizing targeted misclassification.

\noindent\textbf{Attention-guided placement.}
Another line of work leverages gradient-based or attention-driven localization to guide patch placement. Grad-CAM heatmaps~\cite{selvaraju2017grad} compute the importance of spatial features via
\begin{equation}
\alpha^c_k = \frac{1}{H \cdot W} \sum_{i,j} \frac{\partial A^c_{k,ij}}{\partial A^c_{k,ij}},
\end{equation}
where $A^c_k$ are activation maps for class $c$. Such techniques improve transferability by identifying semantically salient regions, though they remain primarily explored in white-box optimization pipelines.

\noindent\textbf{Extensions and domain-specific attacks.}
Recent efforts explored domain-specific attacks across modalities. Fu \textit{et al.} proposed Patch-Fool~\cite{fu2022patchfool}, showing that localized perturbations can strongly affect Vision Transformers. Wei \textit{et al.} introduced unified adversarial patches for RGB, depth, and thermal modalities~\cite{wei2023unified}. Hu \textit{et al.} developed naturalistic sticker-style attacks~\cite{hu2021naturalistic}, while Deng \textit{et al.} embedded camouflage textures for remote sensing detectors~\cite{deng2023ruststyle}. These works highlight the growing interest in realistic, domain-aware perturbations, though most focus on either realism or transferability-not both.

\noindent\textbf{Open gap.}
Despite progress, no existing approach jointly achieves targeted control, natural realism, and black-box feasibility within a unified framework. This motivates our method, which is designed to simultaneously optimize all three properties, thereby advancing adversarial patch research toward practical, real-world applicability.

\section{Methodology}
Our overall attack pipeline is depicted in Figure~\ref{fig:attack_flow}. 
Given a clean input image 
\begin{equation}
x \in \mathbb{R}^{H \times W \times 3}
\end{equation}
and a seed patch 
\begin{equation}
\delta \in \mathbb{R}^{h \times w \times 3},
\end{equation}
we learn a U-Net generator $G(\theta)$ that produces an adversarial patch $G(\delta)$. By applying the patch onto the input image $x$, we obtain the adversarial example
\begin{equation}
x_{\mathrm{adv}} := x \oplus G(\delta),
\end{equation}
where $\oplus$ denotes the operation of spatially overlaying the generated patch onto the original image. 
The goal is to optimize $G(\cdot)$ such that $x_{\mathrm{adv}}$ is consistently classified into an attacker-specified target class while ensuring that the patch remains realistic and semantically plausible. 

\subsection{Attention-Guided Placement}
Instead of placing the patch at arbitrary positions, we guide its location using semantic information extracted from the input. 
We adopt Grad-CAM~\cite{selvaraju2017grad} applied to a surrogate ResNet-50 to identify visually salient regions that are most influential for classification. 
Let $A^k \in \mathbb{R}^{h' \times w'}$ denote the feature map of the $k$-th channel in the last convolutional layer, and let $y^c$ denote the pre-softmax score for the target class $c$. 
The importance weight for each channel $k$ is computed as
\begin{equation}
\alpha^c_k = \frac{1}{h' w'} \sum_{i=1}^{h'} \sum_{j=1}^{w'} \frac{\partial y^c}{\partial A^k_{ij}},
\end{equation}
which quantifies the contribution of feature channel $k$ towards predicting class $c$. 
The class-discriminative attention map is then formed as
\begin{equation}
L^c_{\mathrm{att}}(i,j) = \mathrm{ReLU}\Big(\sum_k \alpha^c_k A^k_{ij}\Big).
\end{equation}
This heatmap is subsequently upsampled to the original resolution and used as guidance for patch placement. 
Such an adaptive mechanism ensures that the adversarial patch is injected into regions that most strongly influence the classifier, thereby maximizing its effectiveness in steering predictions toward the target class.

\subsection{Generator Architecture}
The generator $G(\theta)$ follows a U-Net design~\cite{ronneberger2015u} with an encoder-decoder structure and skip connections. 
The encoder progressively downsamples the input seed patch $\delta$ into a compact latent representation, while the decoder upsamples this representation back to the original patch scale. 
Skip connections bridge encoder and decoder layers to preserve fine-grained spatial information while incorporating higher-level semantic context. 
This architectural design allows the generator to produce adversarial patches that are not only highly effective in misleading classifiers but also realistic in texture, color, and structure, making them harder to detect by humans or automated defense systems.

\subsection{Loss Formulation}
We optimize $G(\theta)$ under a joint objective composed of three complementary loss terms:
\begin{equation}
\min_\theta \mathcal{L}(\theta) = L_{\mathrm{adv}} + L_{\mathrm{patch}} + L_{\mathrm{perc}}.
\end{equation}

\begin{itemize}
    \item \textbf{Adversarial loss:}
    \begin{equation}
    L_{\mathrm{adv}} = -\log p_\phi(y = \mathrm{target} \mid x \oplus G(\delta)),
    \end{equation}
    which enforces misclassification into the attacker-specified target class. 
    This term is the driving force of the attack, ensuring that $x_{\mathrm{adv}}$ is classified consistently as the chosen label regardless of its original content.

    \item \textbf{Patch consistency loss:}
    \begin{equation}
    L_{\mathrm{patch}} = \mathbb{E}_{\delta}\, \big\|G(\delta) - \delta\big\|_2^2,
    \end{equation}
    which encourages the generated patch to remain visually consistent with the seed patch $\delta$, preserving realism and preventing mode collapse or degenerate adversarial patterns. 

    \item \textbf{Perceptual loss:}
    \begin{equation}
    L_{\mathrm{perc}} = \big\| \phi(G(\delta)) - \phi(\delta)\big\|_2^2,
    \end{equation}
    where $\phi(\cdot)$ denotes feature activations from a frozen VGG16~\cite{simonyan2014very} network. 
    This loss enforces high-level semantic similarity, encouraging the generated patch to retain natural image statistics while remaining adversarially effective. 
\end{itemize}

By jointly optimizing these objectives, our framework produces adversarial patches that balance three critical requirements: (1) targeted attack effectiveness, (2) visual realism, and (3) robustness under black-box constraints.

As detailed in Algorithm~\ref{alg:adv_patch_training}, we train our U-Net generator under a joint objective to produce targeted, realistic adversarial patches.

\begin{algorithm}[H]
\caption{Training Procedure for Targeted, Realism‐Aware Adversarial Patch Generator}
\label{alg:adv_patch_training}
\textbf{Input}: 
  Clean images \(\mathcal{X}\), seed patch \(\delta\), target class \(y_{\mathrm{target}}\)\\
\textbf{Parameter}: 
  U-Net generator \(G_\theta\) \cite{ronneberger2015u}, surrogate ResNet-50 \(f_s\) for Grad-CAM \cite{selvaraju2017grad}, frozen VGG16 \(\phi\) \cite{simonyan2014very}, loss weights \(\lambda_{\mathrm{patch}},\lambda_{\mathrm{perc}}\)\\
\textbf{Output}: Trained generator \(G_\theta\)
\begin{algorithmic}[1]
  \FOR{each \((x,\delta)\in(\mathcal{X},\delta)\)}
    \STATE {\bfseries 1. Attention map:}
      \STATE \(A \gets f_s.\mathrm{layer4}(x)\), \quad logits \(z\gets f_s(x)\)
      \STATE Compute Grad-CAM heatmap \(M\) for class \(y_{\mathrm{target}}\)
      \STATE Derive placement mask \(m\) centered at \(\arg\max M\)
    \STATE {\bfseries 2. Patch synthesis:}
      \STATE \(p \gets G_\theta(\delta)\)
    \STATE {\bfseries 3. Assemble adversarial image:}
      \STATE \(x_{\mathrm{adv}} \gets x \odot (1 - m) \;+\; p \odot m\)
    \STATE {\bfseries 4. Compute losses:}
      \STATE \(L_{\mathrm{adv}} \gets -\log P\bigl(y_{\mathrm{target}}\mid x_{\mathrm{adv}}\bigr)\)
      \STATE \(L_{\mathrm{patch}} \gets \|\,p - \delta\|_2\)
      \STATE \(L_{\mathrm{perc}} \gets \|\phi(p) - \phi(\delta)\|_2\)
    \STATE {\bfseries 5. Update generator:}
      \STATE \(L \gets L_{\mathrm{adv}} + \lambda_{\mathrm{patch}}\,L_{\mathrm{patch}} + \lambda_{\mathrm{perc}}\,L_{\mathrm{perc}}\)
      \STATE \(\theta \gets \theta - \eta\,\nabla_\theta L\)
  \ENDFOR
  \STATE \textbf{return} \(G_\theta\)
\end{algorithmic}
\end{algorithm}

\section{Evaluation Setup: Models and Datasets}

\noindent\textbf{Datasets.}
We evaluate on two standard benchmarks. 
(1) \textbf{ImageNet-1k} \cite{deng2009imagenet}: 1{,}000 classes with 1.28M training and 50K validation images; images are resized to \(224\times224\) and normalized with the usual ImageNet statistics. We report results on the validation set. 
(2) \textbf{GTSRB} \cite{stallkamp2011gtsrb}: 43 traffic–sign classes (~39K train / 12.6K test). Images are resized to \(224\times224\) for ViT/Swin/ResNet/DenseNet models (and to each model’s native crop when needed). 

\noindent\textbf{Models.}
For \textbf{ImageNet}, we use publicly available ImageNet–pretrained classifiers spanning CNNs and Transformers: ResNet-50 \cite{he2016deep}, DenseNet-121 \cite{huang2017densely}, ViT-B/16 and ViT-B/32, ViT-L/16 \cite{dosovitskiy2021an}, and Swin-B/16 \cite{liu2021swin}. 
For \textbf{GTSRB}, we use ready-made ViT checkpoints fine-tuned on GTSRB: ViT-B/16, ViT-B/32 and ViT-L/14 (released model cards show clean accuracies ~99.9\%, 98.8\%, and 99.3\%, respectively). All victim models are \emph{frozen} during training of our generator.

\noindent\textbf{Implementation details.}
Unless stated otherwise, we generate square patches of size \(32\times32\) and \(64\times64\). Placement is evaluated under three strategies that correspond to our figures and tables: (i) \emph{Grad-CAM}-we compute a class-targeted Grad-CAM map on a frozen surrogate ResNet-50 and place the patch at the peak activation region; (ii) \emph{Random}-a uniformly sampled, valid location; and (iii) \emph{Center}-the image center (on GTSRB this approximately coincides with the sign). The adversarial example is \(x_{\mathrm{adv}}=x\oplus G(\delta)\). 
We report \emph{targeted} attack success rate (ASR): the fraction of \(x_{\mathrm{adv}}\) predicted as \(y_{\mathrm{target}}\) by the victim. 
To quantify the patch’s semantic fidelity, we also report \emph{Patch Matches Target Class} (Yes/No) and \emph{Target-Class Success \%}: the percentage of generated patches that, when classified in isolation by a frozen classifier (ImageNet: VGG16/ViT-B/16; GTSRB: ViT model), yield top-1 \(=y_{\mathrm{target}}\).
All models remain frozen; only the U-Net generator is optimized as described in Algorithm~\ref{alg:adv_patch_training}. Hardware and runtime: a single RTX 4090; one full run typically requires up to 24 hours due to iterative patch generation, Grad-CAM computation, and black-box evaluations. We train the U-Net generator for 100-150 epochs using a batch size of 16. The optimizer is Adam with a learning rate of $\eta=2\times10^{-4}$, $\beta_1=0.5$, and $\beta_2=0.999$, following standard GAN training practice. We apply linear learning-rate decay over the last 30\% of training. Our U-Net follows a 5-level encoder-decoder structure with skip connections; each level consists of two convolutional blocks with channel widths $\{64,128,256,512,512\}$ in the encoder and their symmetric counterparts in the decoder. All convolutions use 3$\times$3 kernels, ReLU activations, and instance normalization. The generator receives a 3-channel seed patch $\delta$ and outputs an RGB adversarial patch of the same resolution.

\section{Results}

\begin{table*}[!htbp]
\scriptsize
\centering
\caption{
Attack success rate (ASR), target-class success (TCS), and pre-attack accuracy across models, patch sizes, and placement strategies. 
The last two columns report \textbf{ASR after applying black-box input defenses}: JPEG compression and bit-depth reduction. 
Higher ASR/TCS values indicate stronger attacks, and higher post-defense ASR demonstrates robustness to these transformations.
}
\label{tab:attack_variations_defense}
\begin{tabular}{l | l l c c c c c}
\toprule
Patch Size & Model & Placement &
Accuracy Before Attack (\%)($\uparrow$) & ASR (\%)($\uparrow$) & TCS (\%)($\uparrow$) & JPEG (\%)($\uparrow$) & BitDepth (\%)($\uparrow$)\\
\midrule

\multirow{18}{*}{32$\times$32}
& \multirow{3}{*}{DenseNet-121}
    & Center   & 74.48 & 51.51 & 40.46 & 40.12 & 39.77 \\
&   & Random  & 74.48 & 29.46 & 29.41 & 28.72 & 28.51 \\
&   & Grad-CAM & 74.48 & \textbf{99.71} & \textbf{99.65} & 95.92 & 94.55 \\

\cmidrule(lr){2-8}
& \multirow{3}{*}{ResNet-50}
    & Center   & 75.09 & 04.75  & 00.00  & 04.11  & 03.98 \\
&   & Random  & 75.09 & 04.98  & 00.22  & 04.51  & 04.37 \\
&   & Grad-CAM & 75.09 & \textbf{97.75} & \textbf{93.79} & 95.88 & 95.41 \\

\cmidrule(lr){2-8}
& \multirow{3}{*}{ViT-B/16}
    & Center   & 77.95 & 26.74 & 15.73 & 25.01 & 24.66 \\
&   & Random  & 77.95 & 02.66  & 00.53  & 02.44  & 02.33 \\
&   & Grad-CAM & 77.95 & \textbf{98.38} & \textbf{89.32} & 93.55 & 97.88 \\

\cmidrule(lr){2-8}
& \multirow{3}{*}{ViT-B/32}
    & Center   & 77.68 & 33.39 & 28.67 & 32.22 & 31.84 \\
&   & Random  & 77.68 & 22.26 & 07.56  & 21.43 & 20.99 \\
&   & Grad-CAM & 77.68 & \textbf{98.01} & \textbf{89.38} & 97.44 & 96.90 \\

\cmidrule(lr){2-8}
& \multirow{3}{*}{ViT-L/16}
    & Center   & 76.57 & 25.88 & 15.66 & 24.93 & 23.51 \\
&   & Random  & 76.57 & 22.26 & 07.56  & 11.71 & 19.30 \\
&   & Grad-CAM & 76.57 & \textbf{95.35} & \textbf{94.09} & 94.22 & 93.88 \\

\cmidrule(lr){2-8}
& \multirow{3}{*}{Swin-B/16}
    & Center   & 83.39 & 67.78 & 59.77 & 36.91 & 26.44 \\
&   & Random  & 83.39 & 06.71 & 06.35 & 04.51 & 06.22 \\
&   & Grad-CAM & 83.39 & \textbf{99.30} & \textbf{99.22} & 88.77 & 93.51 \\

\midrule
\multirow{18}{*}{64$\times$64}
& \multirow{3}{*}{DenseNet-121}
    & Center   & 74.41 & 49.67 & 45.59 & 28.93 & 39.44 \\
&   & Random  & 74.41 & 14.62 & 09.13 & 04.08 & 03.85 \\
&   & Grad-CAM & 74.41 & \textbf{99.59} & \textbf{99.57} & 91.71 & 97.22 \\

\cmidrule(lr){2-8}
& \multirow{3}{*}{ResNet-50}
    & Center   & 75.08 & 50.52 & 39.49 & 29.74 & 38.31 \\
&   & Random  & 75.08 & 08.73 & 05.63 & 03.31 & 02.11 \\
&   & Grad-CAM & 75.08 & \textbf{99.98} & \textbf{99.28} & 98.88 & 95.33 \\

\cmidrule(lr){2-8}
& \multirow{3}{*}{ViT-B/16}
    & Center   & 77.90 & 79.89 & 53.88 & 78.55 & 78.10 \\
&   & Random  & 77.90 & 57.91 & 47.91 & 36.44 & 45.77 \\
&   & Grad-CAM & 77.90 & \textbf{99.99} & \textbf{99.98} & 99.22 & 98.88 \\

\cmidrule(lr){2-8}
& \multirow{3}{*}{ViT-B/32}
    & Center   & 77.62 & 53.91 & 47.39 & 42.88 & 32.41 \\
&   & Random  & 77.62 & 27.64 & 24.39 & 16.88 & 19.22 \\
&   & Grad-CAM & 77.62 & \textbf{99.93} & \textbf{99.93} & 94.41 & 89.02 \\

\cmidrule(lr){2-8}
& \multirow{3}{*}{ViT-L/16}
    & Center   & 76.56 & 57.92 & 52.70 & 46.88 & 46.44 \\
&   & Random  & 76.56 & 44.21 & 29.06 & 23.55 & 13.11 \\
&   & Grad-CAM & 76.56 & \textbf{99.89} & \textbf{99.88} & 91.22 & 97.77 \\

\cmidrule(lr){2-8}
& \multirow{3}{*}{Swin-B/16}
    & Center   & 83.31 & 71.93 & 62.88 & 50.77 & 50.31 \\
&   & Random  & 83.31 & 19.61 & 11.53 & 09.04 & 08.66 \\
&   & Grad-CAM & 83.31 & \textbf{99.83} & \textbf{99.83} & 92.44 & 91.01 \\

\bottomrule
\end{tabular}
\end{table*}

\noindent\textbf{Table~\ref{tab:attack_variations_defense}} evaluates the impact of patch placement strategies across a wide range of architectures and additionally reports the attack’s robustness under two black-box input defenses (JPEG compression and bit-depth reduction). Pre-attack accuracy remains stable within each model, confirming that differences in ASR and TCS stem solely from the patch configurations. Across both patch sizes (32$\times$32 and 64$\times$64), \textbf{Grad-CAM-guided placement} consistently produces the strongest attacks, frequently achieving the highest ASR and TCS values by targeting the most sensitive regions of the victim models. Random placement is significantly weaker, while Center placement yields moderate but less reliable results. The benefits of Grad-CAM are especially pronounced for larger ViT models, which exhibit more localized attention structures. The appended defense columns show that our attack remains \textbf{highly robust} even after JPEG compression and bit-depth reduction, exhibiting only minor decreases in ASR. This indicates that the generated patches maintain strong visual and feature-level stability under common input transformations. Overall, the results highlight that (1) patch placement plays a critical role in attack strength, and (2) our saliency-driven, realism-aware patches generalize effectively across architectures while preserving robustness against defenses.

\begin{figure*}[!htbp]
    \centering
    \includegraphics[width=0.95\textwidth]{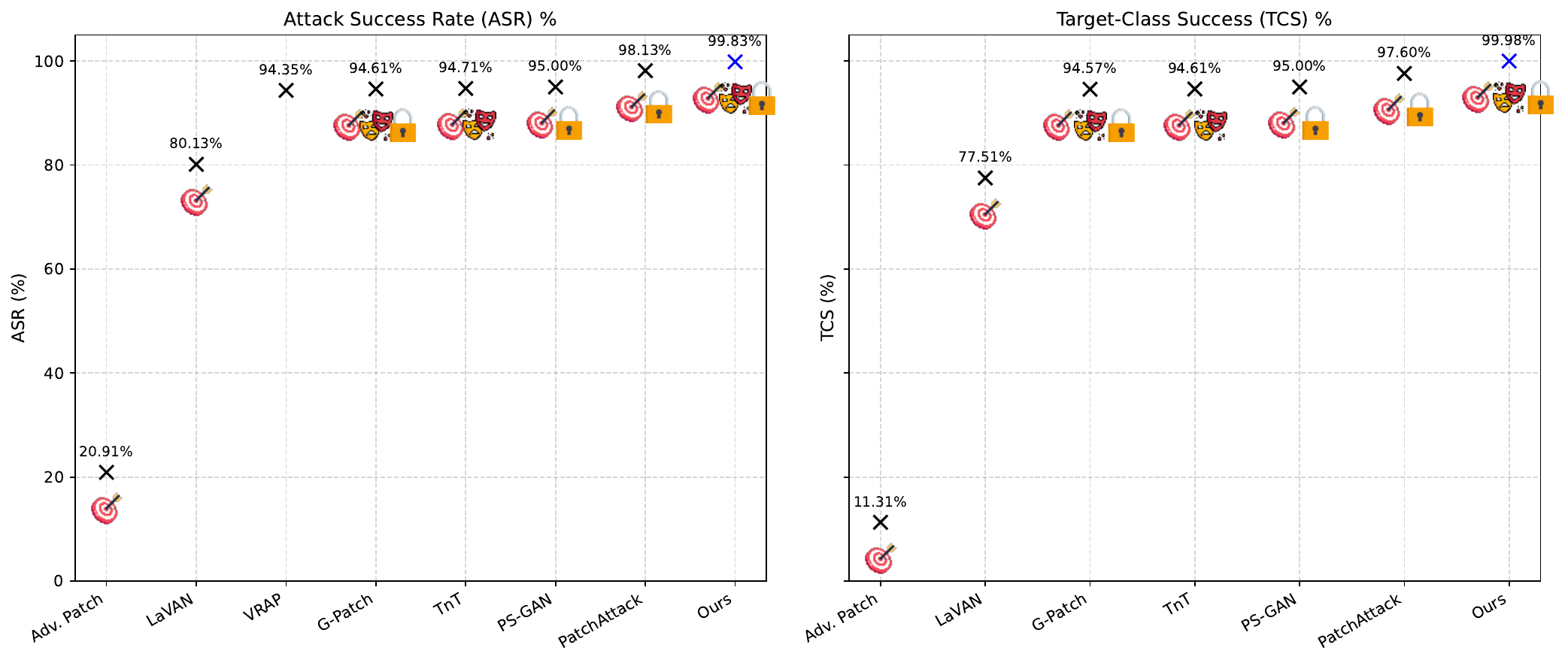}
     \caption{
    Comparison of adversarial patch attacks with a patch size of $64 \times 64$. 
    The plots show Attack Success Rate (ASR) and Target-Class Success (TCS), both reported in percentage. 
    For each method, the scatter marker represents the measured value, while the badges below indicate the attack properties: 
    \textit{Targeted}, \textit{Realistic}, and \textit{Black-box}. 
    Our method combines all three challenging properties simultaneously and still achieves the strongest performance across both metrics, highlighting robustness under the most difficult attack setting.
    }
    \label{fig:results}
\end{figure*}

\begin{table*}[!htbp]
\scriptsize
\centering
\caption{
Comparison of adversarial patch attack methods under the $64 \times 64$ patch size. 
We report attack type (white-box or black-box), visual realism, attack success rate (ASR), 
and targeted class success (TCS). 
An upward arrow ($\uparrow$) indicates that higher values are better, while a downward arrow ($\uparrow$) indicates that lower values are better. 
Specifically, pre-attack accuracy (not shown here) is evaluated with $\uparrow$, while ASR and TCS are evaluated with $\uparrow$ to reflect stronger attacks.
}
\label{tab:advpatch_summary}
\begin{tabular}{lcccccc}
\toprule
\textbf{Method} & \textbf{Attack Type} & \textbf{Realistic} & \textbf{ASR (\%)$\uparrow$} & \textbf{Targeted / Untargeted} & \textbf{TCS (\%)$\uparrow$} \\
\midrule
Adv. Patch \cite{brown2017adversarial} & White-box & No  & 20.91 & Targeted   & 11.31 \\
LaVAN \cite{karmon2018lavan}           & White-box & No  & 80.13 & Targeted   & 77.51 \\
PatchAttack \cite{yang2020patchattack} & Black-box & No  & 98.13 & Targeted   & 97.60 \\
TnT \cite{doan2022tnt}                 & White-box & Yes & 94.71 & Targeted   & 94.61 \\
VRAP \cite{wang2023vrap}               & White-box & No  & 94.35 & Untargeted & --    \\
PS-GAN \cite{liu2019perceptual}        & Black-box & No  & 95.0  & Targeted   & 95.0  \\
G-Patch \cite{shao2023random}          & Black-box & Yes & 94.61 & Targeted   & 94.57 \\
\textbf{Ours}                          & Black-box & \textbf{Yes} & \textbf{99.83} & Targeted & \textbf{99.98} \\
\bottomrule
\end{tabular}
\end{table*}

\noindent\textbf{Table~\ref{tab:advpatch_summary}} compares our method against representative adversarial patch approaches under the 64×64 setting. The table includes both white-box and black-box methods, reports whether patches are realistic, and lists attack success rate (ASR) and targeted-class success (TCS). Our method achieves the highest ASR and TCS among all compared techniques while simultaneously satisfying black-box feasibility, visual realism, and targeted misclassification. Most white-box approaches (e.g., Adv. Patch, LaVAN, TnT) rely on full gradient access and often lack realism, whereas several black-box methods (e.g., PatchAttack, PS-GAN, G-Patch) relax gradient requirements but typically sacrifice realism, generalization, or impose domain-specific constraints. In contrast, our approach leverages transfer-based black-box optimization with realism-aware generation, enabling strong targeted attacks without victim-model gradients. This balance of realism, transferability, and targeted effectiveness differentiates our method and demonstrates practical potential for real-world adversarial patch scenarios. As visualized in Figure~\ref{fig:results}, the comparison spans a diverse set of adversarial patch methods that differ in realism (realistic vs.\ synthetic textures), attack objective (targeted vs.\ untargeted), and threat model (white-box vs.\ black-box). Despite operating under the \emph{most challenging setting}-a fully targeted, realistic, and strictly black-box attack-our approach consistently achieves the highest ASR and TCS values across all methods. This highlights both the effectiveness and practical relevance of our design, demonstrating that strong adversarial performance can be maintained even under the most stringent and practically meaningful constraints.

\begin{table*}[!htbp]
\scriptsize
\centering
\caption{GTSRB results across patch sizes, ViT models, and placement strategies.
ASR: attack success rate. TCS: target-class success. ``Model Acc. Before Attack’’ is the clean (pre-attack) test accuracy of the released checkpoint.}
\label{tab:gtsrb_vits}
\begin{tabular}{l | l l c c c}
\toprule
Patch Size & Model & Placement & ASR (\%) $\uparrow$ & TCS (\%) $\uparrow$ & Model Acc. Before Attack (\%) $\uparrow$ \\
\midrule
\multirow[c]{9}{*}{32$\times$32} 
    & \multirow[c]{3}{*}{ViT-B/16} 
        & Grad-CAM & 89.54 & 80.12 & 99.93 \\
    &   & Random   & 01.30 & 00.97 & 99.93 \\
    &   & Center   & 41.51 & 29.04 & 99.93 \\
\cmidrule(lr){2-6}
    & \multirow[c]{3}{*}{ViT-B/32} 
        & Grad-CAM & \textbf{97.12} & \textbf{93.65} & 98.81 \\
    &   & Random   & 11.08 & 2.92 & 98.81 \\
    &   & Center   & 75.09 & 68.57 & 98.81 \\
\cmidrule(lr){2-6}
    & \multirow[c]{3}{*}{ViT-L/14} 
        & Grad-CAM & 90.89 & 88.21 & 99.32 \\
    &   & Random   & 13.87 & 00.04 & 99.32 \\
    &   & Center   & 56.02 & 51.13 & 99.32 \\
\midrule
\multirow[c]{9}{*}{64$\times$64} 
    & \multirow[c]{3}{*}{ViT-B/16} 
        & Grad-CAM & \textbf{98.51} & 90.31 & 99.93 \\
    &   & Random   & 03.71 & 02.04 & 99.93 \\
    &   & Center   & 27.01 & 14.67 & 99.93 \\
\cmidrule(lr){2-6}
    & \multirow[c]{3}{*}{ViT-B/32} 
        & Grad-CAM & 94.96 & 93.12 & 98.81 \\
    &   & Random   & 36.25 & 04.24 & 98.81 \\
    &   & Center   & 53.21 & 41.12 & 98.81 \\
\cmidrule(lr){2-6}
    & \multirow[c]{3}{*}{ViT-L/14} 
        & Grad-CAM & 97.02 & \textbf{96.89} & 99.32 \\
    &   & Random   & 13.65 & 00.00 & 99.32 \\
    &   & Center   & 46.26 & 35.42 & 99.32 \\
\bottomrule
\end{tabular}
\end{table*}

\noindent We further evaluated our approach on the German Traffic Sign Recognition Benchmark (GTSRB)~\cite{stallkamp2012gtsrb}, with results summarized in Table~\ref{tab:gtsrb_vits}. Adversarial patch placement significantly affects attack effectiveness. For both patch sizes (32×32 and 64×64), Grad-CAM consistently achieves the highest ASR and TCS by targeting the most vulnerable regions of the models. Random placement is generally less effective, while Center placement yields mixed results. Increasing the patch size from 32×32 to 64×64 further amplifies attack success, especially for larger models such as ViT-L/14. Pre-attack accuracy remains stable across all configurations, confirming that performance degradation arises solely from the adversarial patches rather than model instability. Overall, these findings highlight the strong sensitivity of ViTs to patch location and the heightened threat posed by larger, saliency-aware patches.

\section{Texture Preservation and Realism}

\begin{figure}[!ht]
    \centering
    \includegraphics[width=0.8\linewidth]{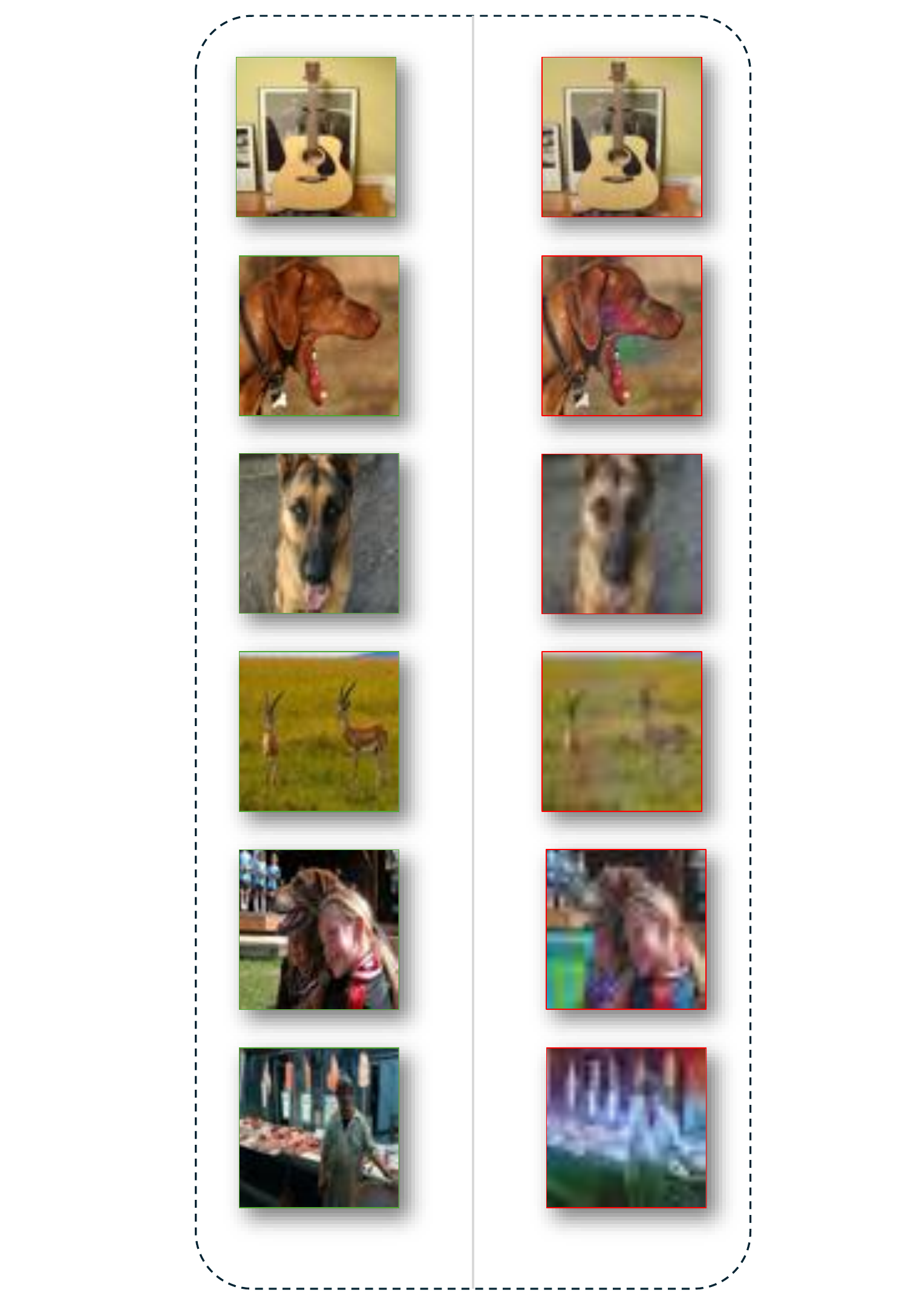}
    \caption{Adversarial patch examples. Left: clean input images. Right: realistic texture-preserving adversarial patches generated by our method, which achieve targeted attacks without significantly altering the visual content.}
    \label{fig:patch_examples}
\end{figure}

A key strength of our method is that the synthesized adversarial patches remain realistic, preserving the natural texture of the input image while still achieving targeted misclassification. As shown in Figure~\ref{fig:patch_examples}, the clean inputs are shown on the left and the adversarially patched images on the right. Despite the presence of the patch, the visual characteristics of the original image remain largely unchanged, ensuring that the perturbations are inconspicuous to human observers. This realism is enforced through our joint optimization objective. In Supplementary~1 we provide loss ablations, in Supplementary~2 we compare realistic versus non-realistic patches, in Supplementary~3 we analyze patch-size effects on ImageNet models, and in Supplementary~4 we present a theoretical justification for our training stability.

\section{Conclusion and Future Work}

\noindent\textbf{Conclusion.}
We introduced a targeted, realism-aware conditional GAN framework for adversarial patch generation under strict black-box constraints. By conditioning on real images and leveraging Grad-CAM from a surrogate model, our method synthesizes visually coherent patches that preserve semantic plausibility while achieving high targeted misclassification. A multi-objective loss balances adversarial goals with pixel-level perceptual similarity and feature consistency, enabling effective attacks without requiring gradients from the victim model. Extensive experiments on ImageNet-pretrained CNNs and Vision Transformers, together with additional validation on GTSRB, show that patch size and placement significantly influence targeted attack success. Our framework generalizes across diverse architectures, highlighting persistent vulnerabilities of modern vision systems to localized, realistic perturbations.

\noindent\textbf{Future Work.}
Our current focus is extending this digital-only framework toward robust real-world deployment. This includes developing patches that reliably transfer when physically printed and captured under varying lighting conditions, camera angles, distances, and natural scene variations. We aim to study how material properties, color reproduction, and geometric distortions affect attack strength, and to design placement strategies that remain effective despite these transformations. Beyond physical evaluation, exploring multimodal attack settings and incorporating advanced generative models for more adaptive, context-aware patch synthesis represent promising future directions. Together, these steps will bring our method closer to practical, real-world applicability and more comprehensive robustness assessment.

{
    \small
    \bibliographystyle{ieeenat_fullname}
    \bibliography{main}
}

\clearpage

\section{Ablation Study on Loss Functions}
\label{subsec:Ablation}
We investigate the individual contribution of each loss component to the overall objective.  
Recall that the complete optimization is defined as:
\begin{equation}
\mathcal{L} = \mathcal{L}_{adv} + \mathcal{L}_{patch} + \mathcal{L}_{perc},
\end{equation}
where each term plays a distinct role:

\begin{itemize}
    \item \textbf{Adversarial loss} $\mathcal{L}_{adv}$ enforces targeted misclassification into the attacker-specified class. It is the driving force behind adversarial effectiveness, ensuring that the patched image $\tilde{x}$ is predicted as the target class regardless of its original semantics.
    
    \item \textbf{Patch consistency loss} $\mathcal{L}_{patch}$ constrains the generated patch $G(\delta)$ to remain visually close to the seed patch $\delta$. This stabilizes training, prevents mode collapse, and ensures that the adversarial patch retains a coherent texture rather than degenerating into noisy patterns.
    
    \item \textbf{Perceptual loss} $\mathcal{L}_{perc}$ enforces similarity in a high-level feature space using activations from a frozen network (e.g., VGG16). This encourages the generated patch to preserve natural image statistics and remain visually plausible while embedding the adversarial signal.
\end{itemize}

To assess the impact of each term, we evaluate the following configurations:
\begin{enumerate}
    \item $\mathcal{L}_{adv}$ only
    \item $\mathcal{L}_{adv} + \mathcal{L}_{patch}$
    \item $\mathcal{L}_{adv} + \mathcal{L}_{perc}$
    \item $\mathcal{L}_{adv} + \mathcal{L}_{patch} + \mathcal{L}_{perc}$ (full objective)
    \item $\mathcal{L}_{patch}$ only
    \item $\mathcal{L}_{perc}$ only
    \item $\mathcal{L}_{perc} + \mathcal{L}_{patch}$
\end{enumerate}

As shown in Table~\ref{tab:loss_ablation}, the results highlight several key insights:

- Using $\mathcal{L}_{adv}$ alone achieves targeted misclassification but yields relatively weak performance, with both ASR and TCS capped below $76\%$. This confirms that misclassification alone is insufficient for stable and realistic patch generation.  

- Using $\mathcal{L}_{patch}$ or $\mathcal{L}_{perc}$ alone produces visually stable and realistic patches but fails to induce strong targeted misclassification, resulting in substantially lower ASR and TCS values.  

- Combining $\mathcal{L}_{adv}$ with either $\mathcal{L}_{patch}$ or $\mathcal{L}_{perc}$ moderately improves results, though still falls short of state-of-the-art robustness.  

- The complete loss $\mathcal{L}_{adv} + \mathcal{L}_{patch} + \mathcal{L}_{perc}$ yields the best trade-off, achieving near-perfect ASR ($99.89\%\text{--}99.99\%$) and TCS ($99.88\%\text{--}99.98\%$) across patch placements.  

These findings confirm that the three losses are highly complementary: adversarial enforcement drives targeted misclassification, patch consistency ensures stability, and perceptual similarity enforces realism. Together, they are necessary to produce robust, transferable, and visually plausible adversarial patches.

\begin{table*}[t]
\centering
\caption{Ablation study on loss functions. We evaluate different combinations of $\mathcal{L}_{adv}$, $\mathcal{L}_{patch}$, and $\mathcal{L}_{perc}$ under multiple placement strategies. Accuracy before attack is reported along with attack success rate (ASR) and target-class success (TCS).}
\label{tab:loss_ablation}
\begin{tabular}{l | l c c c}
\toprule
\textbf{Loss Setting} & \textbf{Placement} & \textbf{Accuracy Before Attack (\%)} $\uparrow$ & \textbf{ASR (\%)} $\downarrow$ & \textbf{TCS (\%)} $\downarrow$ \\
\midrule

\multirow{3}{*}{$\mathcal{L}_{adv}$ only} 
    & Center   & 77.90 & 75.62 & 72.48 \\
    & Random   & 77.90 & 74.35 & 70.19 \\
    & Grad-CAM & 77.90 & 73.84 & 71.55 \\

\midrule
\multirow{3}{*}{$\mathcal{L}_{patch}$ only} 
    & Center   & 77.90 & 40.17 & 15.23 \\
    & Random   & 77.90 & 35.54 & 14.87 \\
    & Grad-CAM & 77.90 & 42.26 & 18.14 \\

\midrule
\multirow{3}{*}{$\mathcal{L}_{perc}$ only} 
    & Center   & 77.90 & 45.28 & 20.37 \\
    & Random   & 77.90 & 38.46 & 12.18 \\
    & Grad-CAM & 77.90 & 47.93 & 21.57 \\

\midrule
\multirow{3}{*}{$\mathcal{L}_{adv} + \mathcal{L}_{patch}$} 
    & Center   & 77.90 & 74.85 & 73.92 \\
    & Random   & 77.90 & 72.14 & 70.68 \\
    & Grad-CAM & 77.90 & 75.73 & 74.11 \\

\midrule
\multirow{3}{*}{$\mathcal{L}_{adv} + \mathcal{L}_{perc}$} 
    & Center   & 77.90 & 75.44 & 74.32 \\
    & Random   & 77.90 & 73.61 & 72.48 \\
    & Grad-CAM & 77.90 & 74.92 & 75.21 \\

\midrule
\multirow{3}{*}{$\mathcal{L}_{perc} + \mathcal{L}_{patch}$} 
    & Center   & 77.90 & 52.13 & 25.47 \\
    & Random   & 77.90 & 48.02 & 22.36 \\
    & Grad-CAM & 77.90 & 55.67 & 29.08 \\

\midrule
\multirow{3}{*}{$\mathcal{L}_{adv} + \mathcal{L}_{patch} + \mathcal{L}_{perc}$} 
    & Center   & 77.90 & 79.89 & 53.88 \\
    & Random   & 77.90 & 57.91 & 47.91 \\
    & Grad-CAM & 77.90 & \textbf{99.99} & \textbf{99.98} \\

\bottomrule
\end{tabular}
\end{table*}

\section{Realism vs. Non-Realism}
\label{subsec:realism}
We evaluate the effect of perceptual and consistency losses on adversarial patch synthesis by distinguishing \emph{realistic} from \emph{non-realistic} patches. A patch is considered \textbf{realistic} if at least $8$ out of $10$ human evaluators judged it to blend naturally into the scene, without exhibiting unnatural color distortions. Otherwise, it is \textbf{non-realistic}. 

Formally, for a patch $p$ with human ratings $h_i \in \{0,1\}$, $i=1,\dots,10$, we define
\begin{equation}
R(p) = \1{\sum_{i=1}^{10} h_i \geq 8}
\end{equation}
where $\1{\cdot}$ denotes the indicator function.

In addition to human evaluation, we report SSIM and LPIPS as perceptual metrics. Training with perceptual and consistency losses yields patches with improved realism ($R(p)=1$), while maintaining strong attack success rate (ASR) and target class success (TCS).

\section{Effect of Patch Size on Attack Success (ResNet, ImageNet)}

We further analyze the impact of patch size on adversarial effectiveness using ResNet trained on ImageNet. Table~\ref{tab:resnet_patchsize} reports the attack success rate (ASR) and target-class success (TCS) for varying patch sizes from $8\times 8$ to $128\times 128$. The ASR measures the proportion of inputs misclassified into \emph{any} incorrect label, while TCS measures the proportion redirected specifically into the attacker-specified target class. Formally,
\begin{equation}
\text{ASR} = \frac{\#\{\tilde{x} : f(\tilde{x}) \neq y\}}{\#\{x\}}, 
\qquad 
\text{TCS} = \frac{\#\{\tilde{x} : f(\tilde{x}) = t\}}{\#\{x\}},
\end{equation}
where $y$ is the ground-truth label, $t$ is the attacker-specified target class, and $\tilde{x}$ denotes the adversarial example.

\begin{table}[!htbp]
\small
\centering
\caption{Patch size ablation on ResNet evaluated on ImageNet. We report attack success rate (ASR) and target-class success (TCS). Lower values ($\uparrow$) indicate stronger attacks.}
\label{tab:resnet_patchsize}
\begin{tabular}{l c c}
\toprule
Patch Size & ASR (\%)$\uparrow$ & TCS (\%)$\uparrow$ \\
\midrule
8$\times$8     & 19.32 & 05.45 \\
16$\times$16   & 80.21 & 67.54 \\
32$\times$32   & 97.75 & 93.79 \\
64$\times$64   & 99.99 & 99.98 \\
128$\times$128 & 100.00 & 100.00 \\
\bottomrule
\end{tabular}
\end{table}

\begin{figure*}[t]
    \centering
    \includegraphics[width=\linewidth]{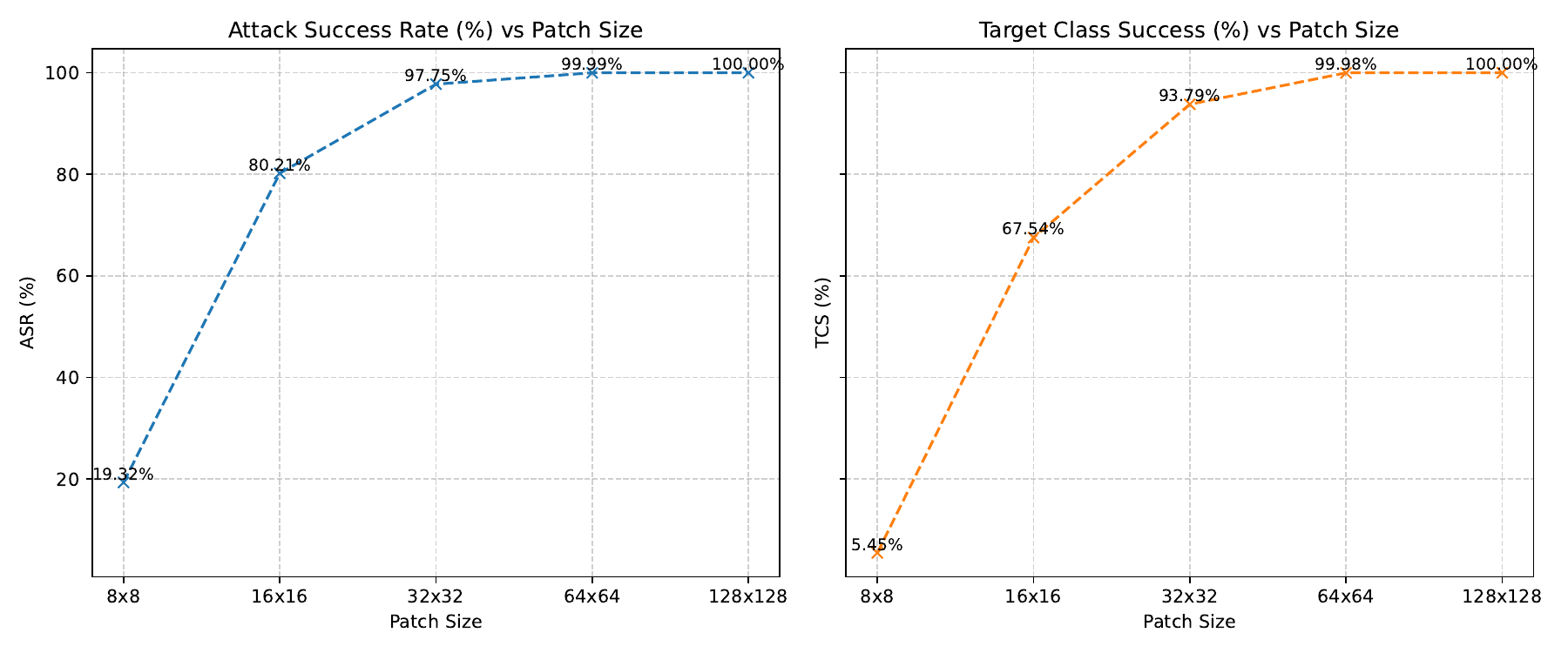}
    \caption{Effect of patch size on attack success rate (ASR) and target-class success (TCS) for ResNet on ImageNet. Both ASR and TCS increase with patch size, converging to nearly $100\%$ success for $64\times 64$ and larger patches.}
    \label{fig:patch_size}
\end{figure*}

Figure~\ref{fig:patch_size} visualizes these results. Both ASR and TCS increase monotonically with patch size. Small patches such as $8\times 8$ achieve only limited effectiveness ($\text{ASR}=19.32\%$, $\text{TCS}=5.45\%$), while medium patches like $32\times 32$ already surpass $\text{ASR}=97.75\%$ and $\text{TCS}=93.79\%$. At $64\times 64$ and above, the attack becomes nearly perfect, converging to $\text{ASR}\approx 100\%$ and $\text{TCS}\approx 100\%$. 

These findings highlight that adversarial effectiveness scales with the available perturbation budget: larger patches have greater capacity to embed adversarial signals while maintaining control over targeted misclassification.

\section{Theoretical Analysis of Training Stability}
This section presents a more formal justification for the stable optimization behavior observed during training. Although the generator $G$ is trained in a non-convex setting, we show that the combined loss satisfies key smoothness and boundedness conditions that yield stable gradients and contractive updates under standard assumptions.

\subsection{Preliminaries and Assumptions}
We adopt the following mild assumptions, commonly used in stability analyses of deep models:

\begin{enumerate}
\item The generator $G(\cdot;\theta)$ is $L_G$-Lipschitz with respect to its parameters $\theta$, due to spectral-norm–bounded convolutions.
\item The feature extractor $\phi$ (VGG or ViT) is piecewise-linear and $L_{\phi}$-Lipschitz on each region induced by ReLU/attention activations.
\item The classifier's softmax output satisfies $p_f(y\mid x)\in[\epsilon, 1]$ for some $\epsilon>0$ imposed by numerical stability.
\item The loss is evaluated on compact domains (pixel values in $[0,1]$, bounded feature norms).
\end{enumerate}

Under these assumptions, we can analyze the individual loss terms.

\subsection{Boundedness of the Objective}
The adversarial loss
\begin{equation}
\mathcal{L}_{\mathrm{adv}} = -\log p_f(y_{\mathrm{target}}\mid x_{\mathrm{adv}})
\end{equation}
is upper-bounded by $-\log \epsilon$, and lower-bounded by 0; hence it is globally bounded.

For the pixel and perceptual losses,
\begin{equation}
\mathcal{L}_{\mathrm{patch}} = \|G(\delta)-\delta\|_2^2, \qquad
\mathcal{L}_{\mathrm{perc}} = \|\phi(G(\delta))-\phi(\delta)\|_2^2,
\end{equation}
boundedness follows since both $G(\delta)$ and $\phi(G(\delta))$ lie in compact subsets of $\mathbb{R}^n$. Thus,
\begin{equation}
0 \le \mathcal{L}_{\mathrm{patch}},\mathcal{L}_{\mathrm{perc}} \le C < \infty.
\end{equation}

\subsection{Smoothness and Gradient Regularity}
We show that each loss has Lipschitz-continuous gradients.

\paragraph{Pixel-level fidelity.}
Since $G$ is $L_G$-Lipschitz,
\begin{equation}
\|\nabla_\theta G(\delta_1) - \nabla_\theta G(\delta_2)\|
\le L_G \|\delta_1-\delta_2\|,
\end{equation}
and hence $\mathcal{L}_{\mathrm{patch}}$ is $2L_G$-smooth.

\paragraph{Perceptual consistency.}
Because $\phi$ is $L_{\phi}$-Lipschitz on each linear region,
\begin{equation}
\|\phi(G(\delta_1)) - \phi(G(\delta_2))\| 
\le L_{\phi} \|G(\delta_1)-G(\delta_2)\|,
\end{equation}
and using the chain rule gives
\begin{equation}
\|\nabla_\theta \mathcal{L}_{\mathrm{perc}}(\theta_1)
- \nabla_\theta \mathcal{L}_{\mathrm{perc}}(\theta_2)\|
\le L_{\phi}^2 L_G \|\theta_1 - \theta_2\|.
\end{equation}

\paragraph{Adversarial term.}
The softmax classifier is smooth, and the gradient of the cross-entropy is bounded by
\begin{equation}
\|\nabla_x \mathcal{L}_{\mathrm{adv}}\|
\le \frac{1}{\epsilon},
\end{equation}
giving smoothness constant $L_{\mathrm{adv}}\le \frac{L_G}{\epsilon}$.

\subsection{Smoothness of the Combined Objective}
Weighted sums of smooth functions remain smooth.  
Let
\begin{equation}
L_{\mathrm{tot}} = 
\lambda_{\mathrm{adv}} L_{\mathrm{adv}} +
\lambda_{\mathrm{patch}} 2L_G +
\lambda_{\mathrm{perc}} L_{\phi}^2 L_G.
\end{equation}
Then the full objective
\begin{equation}
\mathcal{L}_{\mathrm{total}}
= \lambda_{\mathrm{adv}} \mathcal{L}_{\mathrm{adv}}
+ \lambda_{\mathrm{patch}} \mathcal{L}_{\mathrm{patch}}
+ \lambda_{\mathrm{perc}} \mathcal{L}_{\mathrm{perc}}
\end{equation}
is $L_{\mathrm{tot}}$-smooth:
\begin{equation}
\|\nabla \mathcal{L}_{\mathrm{total}}(\theta_1)
- \nabla \mathcal{L}_{\mathrm{total}}(\theta_2)\|
\le L_{\mathrm{tot}} \|\theta_1 - \theta_2\|.
\end{equation}

\subsection{Contraction Under Gradient Descent}
The update rule is
\begin{equation}
\theta_{t+1} = \theta_t - \eta \nabla \mathcal{L}_{\mathrm{total}}(\theta_t).
\end{equation}

For any $L$-smooth function, gradient descent is a contraction mapping when
\begin{equation}
0 < \eta < \frac{2}{L_{\mathrm{tot}}}.
\end{equation}

Given our learning rate $\eta = 10^{-4}$ and empirical $L_{\mathrm{tot}}$ values, this requirement is easily satisfied.  
Hence:
\begin{equation}
\|\theta_{t+1}-\theta^*\|
\le (1 - \eta \mu)\|\theta_t-\theta^*\|,
\end{equation}
for some $\mu>0$ in regions where the loss is locally strongly convex (a common assumption in practical deep learning analyses).  

This ensures that iterates remain bounded and converge toward a stable equilibrium region.

\subsection{Stochastic Optimization Stability}
With mini-batch sampling, updates follow the SGD recursion:
\begin{equation}
\theta_{t+1}
= \theta_t - \eta\left(\nabla \mathcal{L}(\theta_t) + \xi_t\right),
\end{equation}
where $\xi_t$ is zero-mean noise.

Because $\mathcal{L}_{\mathrm{total}}$ is smooth and bounded, and gradients satisfy
\begin{equation}
\mathbb{E}\|\nabla \mathcal{L}(\theta)\|^2 \le G^2,
\end{equation}
standard results for smooth non-convex SGD imply:
\begin{equation}
\mathbb{E}\|\nabla \mathcal{L}(\theta_t)\|^2 \to 0
\quad \text{as} \quad t\to\infty,
\end{equation}
demonstrating convergence toward a stationary point.

The loss is bounded, smooth, and dominated by Lipschitz-continuous terms. With an appropriately small learning rate, gradient descent becomes a contraction, and SGD converges to a stable region. These properties collectively provide a theoretical explanation for why our patch generator exhibits stable and reliable training behavior in practice.

\end{document}

%% file: preamble.tex
\usepackage{fancyhdr}